%% file: main.tex
\documentclass{WileyASNA-v1}

\articletype{ORIGINAL ARTICLE}

\received{}
\revised{}
\accepted{}

\usepackage{hyperref}
\usepackage{graphics}

\begin{document}
	\title{A study of Neural networks point source extraction on simulated Fermi/LAT Telescope images.} 
    \author[1,4]{Drozdova Mariia}
    \author[2]{Broilovskiy Anton}
    \author[2]{Ustyuzhanin Andrey}
    \author[3]{Malyshev Denys}
	
	
    \address[1]{Department of General and Applied Physics, Moscow Institute of Physics and Technology, Moscow, Russia}
    \address[2]{Department of Applied Mathematics and Computer Science, National Research University Higher School of Economics, Moscow, Russia}
    \address[3]{Institut f{\"u}r Astronomie und Astrophysik T{\"u}bingen, Universit{\"a}t T{\"u}bingen, Sand 1, D-72076 T{\"u}bingen, Germany}
    \address[4]{Ecole Polytechnique, Palaiseau, France}
    
\corres{Mariia Drozdova, Ecole Polytechnique, Palaiseau, France. \newline \email{maria.drozdova@polytechnique.edu}}

\abstract{Astrophysical images in the GeV band are challenging to analyze due to the strong contribution of the background and foreground astrophysical diffuse emission and relatively broad point spread function of modern space-based instruments.
In certain cases, even finding of point sources on the image becomes a non-trivial task. We present a method for point sources extraction using a convolution neural network (CNN) trained on our own artificial data set which imitates images from the Fermi Large Area Telescope. These images are raw count photon maps of $10\times 10$ deg$^{2}$ covering energies from 1 to 10 GeV. We compare different CNN architectures that demonstrate accuracy increase by $\approx 15 \%$ and reduces the inference time by at least the factor of 4 accuracy improvement with respect to a similar state of the art models.}





\maketitle
\section{Introduction}

Modern gamma-ray telescopes, e.g. Fermi/LAT~\cite{lat} or AGILE~\cite{agile} explore the sky in 0.1 -- 100~GeV energy band.
These instruments are characterised by a broad field of view ($20\%$ of all-sky for Fermi/LAT) and usually operate in ``all sky-survey'' mode in which the telescope continuously scans the sky.

Several different-class objects can appear on a typical GeV map -- point-like galactic and extragalactic (isotropic) sources (e.g. pulsars and active galactic nuclei); extended galactic and extragalactic sources (e.g. supernova remnants and jets from nearby radio-galaxies); galactic (mainly hadronic emission from gas clouds; variable at a degree-scale, see, e.g.~\citet{fermi_galdiffuse}) and extragalactic diffuse emission. Characteristic angular sizes of extended sources vary from less than a degree to comparable to all-sky scales (e.g. Fermi bubbles). A finite and energy-dependent ($\sim 1^\circ$ at 1~GeV and $0.1^\circ$ at 100~GeV, see e.g.~\cite{lat_psf}) resolution of the photon arrival direction reconstruction leads to the additional artificial broadening even of point-like sources. Due to this broadening extended and even point-like sources can be confused with the similar angular-scale features in diffuse galactic emission. Additional complexity arises from the low-statistics of the observed data, which becomes thus a subject of Poisson fluctuations.

In this paper, we propose a point source extraction technique which allows to detect and separate point-like sources from the diffuse component returning its position with image pixel resolution.

Since \cite{AlexNET}, neural networks have been invading different areas of research from medical image analysis by \cite{medical_nn} to jet structure analysis in particle physics by \cite{deepadvertising}. Concerning approximation of the real data by our artificial dataset we assume that our neural networks do not require for operation any additional information.  Most state-of-the-art point source extraction algorithms in astrophysics imply specific parameters that should be tuned for the particular dataset characteristics, reducing, hence, the generality of the approach. The examples can be found in Sec.~\ref{sec:related_work}.
\newpage

The point source extraction problem can be reduced to a segmentation problem with two classes: background and sources. We address the problem as a segmentation problem via Unet by \cite{unet} and via our neural net architecture. The main problem is the class imbalance as the number of background pixels four orders of magnitude times larger than the signal pixel number.

We generate our artificial dataset and define metrics to evaluate our results.  We compare these methods with classical approaches SExtractor by \cite{sxtractor1996} and D3PO by \cite{d3po} and object detection approach via YOLO by \cite{yolo}, and estimate their performance.

This article consists of seven parts. In the second part, we will analyze the literature. The third part reveals the formulation of the problem with the dataset and metric description; the fourth describes models, preprocessing methods, and training process -- the quality of the models and error research in the fifth part. The sixth part is devoted to the result discussion.

\section{Related Work}
\label{sec:related_work}
There are many algorithms for solving the point source extraction problem. In the optics range, one of the first algorithms is CLEAN by \cite{clean}. It works in Fourier inverse image space and removes source by source. One of the most well-known heuristic methods is SExtractor by \cite{sxtractor1996}. This algorithm analyses an astrophysical image in the following stages: a) the algorithm estimates a background level and constructs the background map, b) detecting object via thresholding, and peak finding methods c) detected objects are checked and separated, if necessary, d) algorithm cleans spurious detections and estimates the total magnitude of found objects e) neural network classifies point-like sources. Finally SExtractor returns only the list of point sources without denoised and deconvolved signal. It also estimates the magnitude (significance) of sources which is not achieved currently by our technique.

Another approach consists of setting up the probabilistic relations between observations and sources. Powellsnakes by \cite{powellsnakes} is an algorithm for detecting objects on multi-frequency astronomical images based on the Bayesian filters that use a prior for position, size, number of sources, flux, and spectral parameters. It uses PSF templates for detection and classification. However, no implementation of this approach is available, so we exclude it from our comparison. Another probabilistic method is D3PO by \cite{d3po} that aims at the concurrent denoising, deconvolution, and decomposition of photon observation into two components: the diffuse component, and point-like photon sources. This algorithm uses Bayesian inference and is based on the difference between the intensity of the source component and the cosmic background. 

D3PO reconstructs the diffused and point-like photon flux from a single photon count image. The method does not distinguish between diffuse sources and background, and both are considered to be the diffuse contribution. The method is based on several predetermined well-known prior distributions. 

Convolution neural networks (CNN) are currently one of the most influential and accurate image processing algorithms available. CNNs are widespread in image processing tasks spanning from the recognition of handwritten digits by \cite{lenet} to the analysis of the complex medical images by \cite{brain}. Recently this approach has been applied to the domain of optical astrophysical analysis.  \cite{fermi_cnn} describes the process of astrophysical image reconstruction using leNet-like architecture. \cite{gan} introduce the method of restoring astrophysical images using Generative Adversarial Networks which enable recovering features from heavily noised images. State-of-the-art deep learning algorithm from image-detection field YOLO (you only look once) \cite{yolo} introduces a deep network which returns bounding (imaginary) boxes for predefined objects during inference. We apply this network for our point-source detection problem with a fixed size of a bounding box. It is discussed further in the Results section.

\section{Problem description}
As noticed above, the images in the GeV band can be affected by the substantial contribution of the diffuse background. In what below we describe the algorithm able to isolate point-like sources in astrophysical maps.

To efficiently train the models, we need plenty of data, and the real data is not enough. In real life, we have just one sky with a finite number of regions/sources. Thus we selected to use the artificial dataset based on point sources' catalogue and state-of-the-art templates for galactic and extragalactic diffuse emissions~\cite{3fgl_catalogue,fermi_galdiffuse}, see Sec.~\ref{sec:dataset} for details. 
It replicates the properties of the signal and background without dependence on specific prior distributions. 

\subsection{Dataset} 
\label{sec:dataset}

The dataset contains simulated images of objects similar to the ones taken by Fermi/LAT, see, e.g.~\cite{3fgl_catalogue,4fgl_catalogue} for catalogues of detected sources. In addition to images, we provide coordinates of all the point sources within those images. Such dataset is used to test point source extraction techniques, and we assume that it is realistic enough to estimate the performance of the considered algorithms as on real data. We do not limit ourselves to high galactic latitudes with low background level; the dataset contains a significant amount of images with substantial galactic diffuse background contamination. 

Fermi/LAT produces images as raw count maps of photons of selected energy band detected during a certain period. We have chosen the energy range of 1--10~GeV (close to optimal Fermi/LAT sensitivity band for point-like sources) and a time span of 10 years (similar to current mission duration $\sim 12$~years). Each image corresponds to a sky segment of $10^\circ\times10^\circ$ degrees or $200\times200$ pixels. Coordinates of point sources refer to a local Cartesian system of an image with centre at its top left corner. We centre the images at random positions in the sky.

To simulate this dataset, we use the Fermi tools software \cite{website:fermi-tools}. The astrophysical Diffusive Background is modelled as a sum of two standard templates (extragalactic: iso\_P8R2\_CLEAN\_v6 and galactic: gll\_iem\_v06.fits)  at corresponding sky segment. We add point sources randomly distributing them on the picture. Their brightness is sampled from the range [0.5*minimal brightness of real 3FGL catalogue sources found in the sky segment from which we retrieved background; 1.15*maximal brightness of these real sources]. The number of the sources per image is sampled from the range [number of these real sources +-2]. Based on this data, an XML file is created and used to generate an image with the help of \textit{gtmodel} routine of Fermi tools. To make the data realistic (only integer number of photons can be observed), we replace the value in each pixel of the produced map with a value derived from a Poisson distribution with a mean equal to the value simulated for the corresponding pixel.

\begin{figure}
    \centering
    \includegraphics[width=\columnwidth]{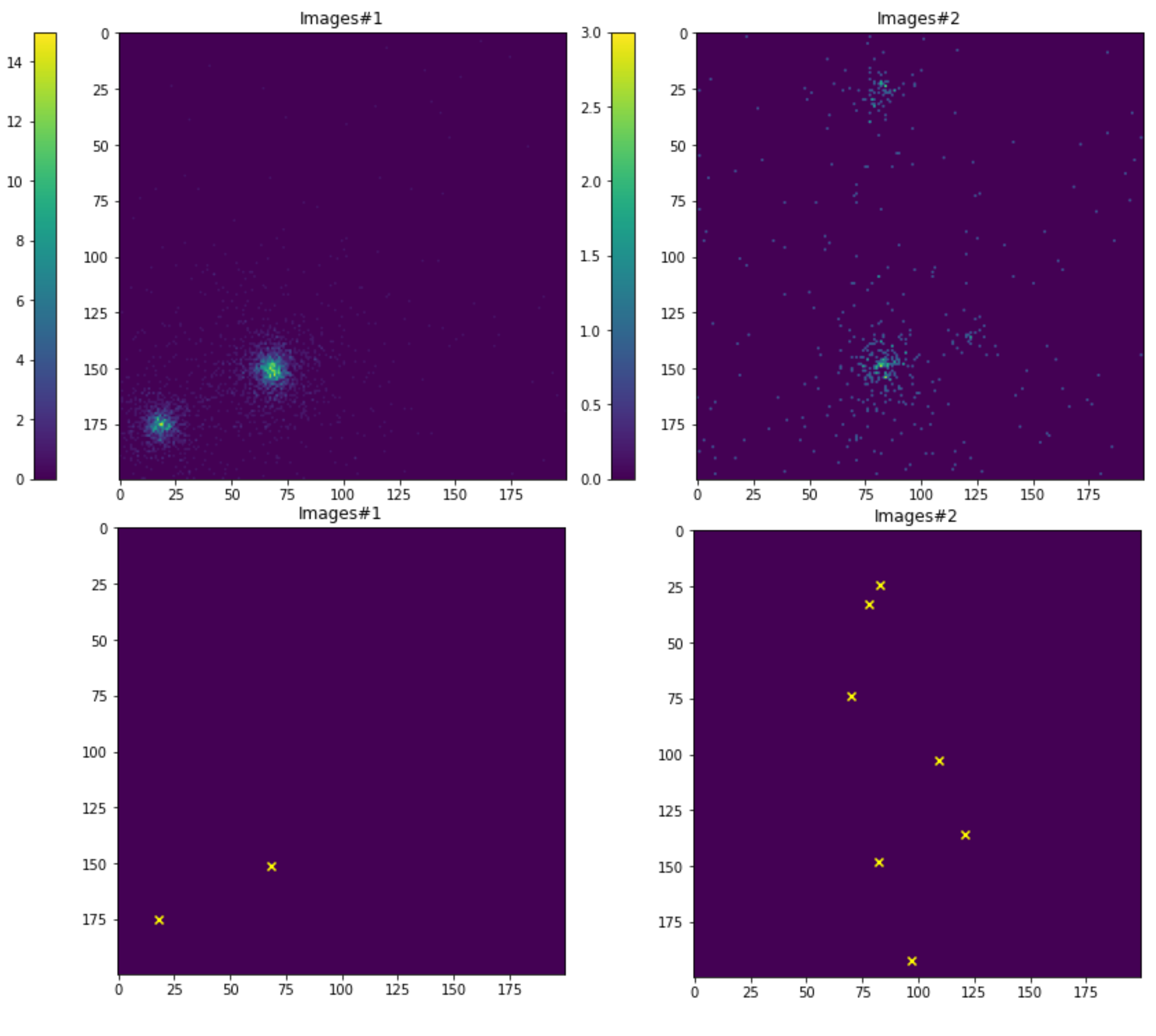}
    \caption{Images$\#$1 is the first type of images where the brightness of the sources is higher than the brightness of the background. Images$\#$2 shows the second type where the brightness of sources and background are comparable. Masks here stand for images where we marked all real point sources as white crosses. }
    \label{fig:dataset_ex}
\end{figure}

There are two kinds of images in the dataset. One of them is where the brightness of some sources is greater than the brightness of the background and the ones where these intensities are comparable, including the ones with a strong background. (Fig. \ref{fig:dataset_ex}).

As far as this dataset has been generated based on the real data, it can be used for comparison of various point source extraction algorithms. The dataset consists of 3692 images. During the training process, 2954 of them was used for training 738 for validating and 407 for testing.
\subsection{Metrics} 
\label{desc:metrics}

To analyze the performance of point extraction models a metric which allows defining the distance between the set of predicted sources (set $A$) and ground truth sources (set $B$) has to be identified. We propose this metric $dist(A,B)$ to be Chamfer Distance, introduced for 3D data analysis (e.g. for 3D Object Reconstruction in \cite{chamferDistance} ):

\begin{align}
\label{eq:metric}
& dist(A, B) = Dist(A, B) + Dist(B, A)  \\ \nonumber
& Dist(A, B) = \sum_{a \in A} \min_{b \in B}\|a - b\|_2    
\end{align}

Here $\|a - b\|_2 = \sqrt{\sum{\|a - b\|^2}}$ stands for a Euclidean norm.

To accelerate its calculations, we used the BallTree structure \cite{balltree}. 
This metric is symmetric, non-negative, continuous, and piece-wise smooth.

Below we consider predicting model $f$ which acts on an image $x$ and returns predicted sources list $f(x)$. Let $y$ be the list of ground truth sources in the corresponding image $x$. In case if $x$ belongs to a collection of images $X$, a corresponding collection $Y: y\in Y$ can be considered. For a given collection of images $X$ and corresponding collection of ground-truth coordinates $Y$ we define the score of the predicting model $f$ as
\begin{equation}
\label{eq:score}
\text{score}(f,X,Y) = \frac{1}{|X|}\sum_{x, y \in X, Y} dist(y, f(x))
\end{equation}

where $|X|$ is the number of images in the considered collection $X$.  

This metric assumes that both predicted and ground truth source sets are not empty. Otherwise,
we introduce penalizing term (\textbf{P}) to the score for each not found or falsely detected source by a fixed constant per point:

$$ P = \frac{1}{8}* H *|C| \text{, where}$$
$C$ is a non-empty set of points, and $H$ is a size in pixels of a side of our square-image.
 \textbf{P} is used as a score when either we have a picture without point sources, but some were predicted, either no point sources were predicted for an image which contains ground-true point sources.
 
 This metric score will be significant in the case when the algorithm fails to predict real sources as such sources do not have close neighbours from predicted ones. Thus, the term corresponding to the distance from such sources will be large. The exception occurs if sources are close to each other. If one of them is detected, the distance for other ones will not be very significant. The score will be large as well in case when the algorithm detects false sources due to the distance from them to closest real sources. If the Chamfer score is small, it means that all real sources have detected ones close to them and all detected sources have real ones close to them.

For a more detailed analysis of errors, we will also consider the following standard metrics: F1 score and True Positive Rate(TPR) \cite{roc}.
We do not consider False Negative Rave and True Negative Rate because the number of negative elements tends to be rather large; therefore, these metrics will not work.

\section{Models} 

We search for a mapping from original astrophysical images to point source coordinates of these images. Chosen Chamfer distance measure the quality of the introduced models.
As baselines, we choose D3PO algorithm and SExtractor. We suggest a simple neural network with four layers, UNet \cite{unet} and FoCNN as our solutions. All of them are convolution neural networks \cite{AlexNET}.

Before describing the architectures, we notice that the dataset is very unbalanced. It consists of a significantly different number of sources points and background points. The mean number of point sources per image is equal to 7 against a mean number of background pixels equal to 39993. Due to class imbalance, the networks tend to output small values even for point sources which means that it tends to favour classifying any pixel as background if the values lower to 0.5 are chosen as a threshold. Thus, we introduce it as a parameter -  a threshold for network output. Pixels of the output image, which have values equal or greater than this threshold we identify as point sources. The value of this parameter is defined through the optimization and is different from 0.5 used in a 2-class classification problem.

\subsection{Images cropping}
We already introduced a threshold as one of the class imbalance technique. The second technique is an image cropping. It also speeds up the learning process. From image $200 \times 200$ we crop image $s \times s$ where $s$ may be various for different network architectures.

\begin{figure}[t]
    \centering
    \includegraphics[width=\columnwidth]{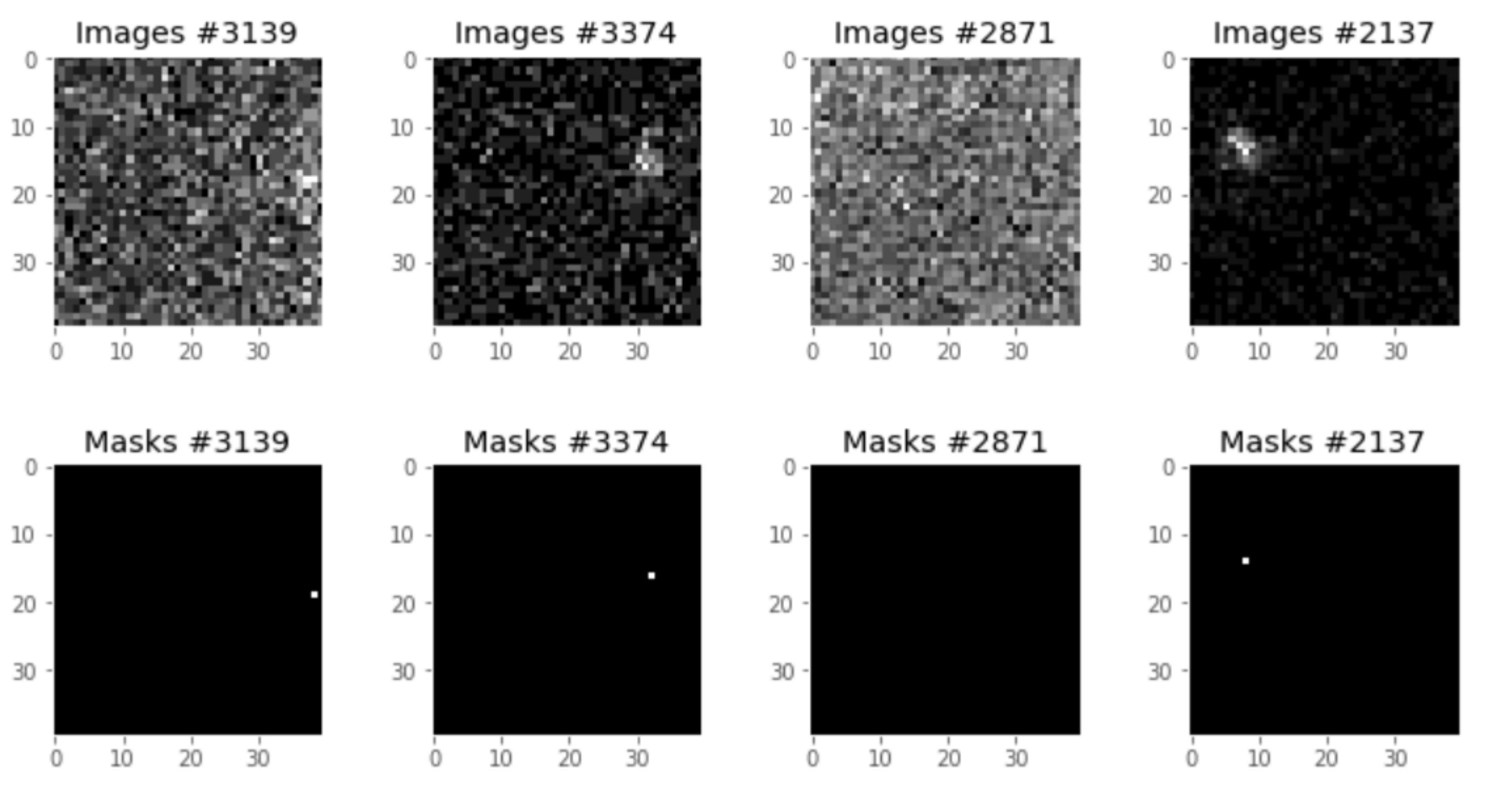}
    \caption{Examples of cropped images and masks with size 40x40 pixels. Images are again raw photon counts. Masks contain only their ground-truth point sources marked as 1-pixel white squares.}
    \label{fig:batch}
\end{figure}

Sources are not uniformly distributed over an image; if we crop randomly, from 2000 crops on average, only 500 will contain sources. But if a neural network during training stage receives empty $s \times s$ it will lead to small training rates. That is why, additionally to the threshold, we implement picking randomly only a fraction of images without sources. Another technique is random sampling. With a chosen probability, we take either a sample with at least one source or a random one. Using these techniques, we can control a number of sources during the training process. In the Figure \ref{fig:batch} we give examples of cropping for $s = 40$.

\subsection{Network Architectures}
\subsubsection{Simple CNN}

Small Convolutional Neural Networks have a significant advantage over large architectures as they contain few parameters and are quickly trained. We use a small CNN with four layers (Table \ref{tab_cnn}).

This network takes an image as an input and returns an output of the same size. The sigmoid function maps output values from 0 to 1, where one corresponds to the presence of the source at the given pixel. To train this network, we optimize Cross-Entropy loss from \cite{Goodfellow2016}. Also, we use re-weighing of the loss function to penalize more for missing the source and less for missing background. 

\begin{center}
\begin{table}
\centering
\makebox[0pt][c]{\parbox{0.9\textwidth}
{
\begin{minipage}[b]{1.0\linewidth}\centering
 \begin{tabular}{|| c | c | c | c | c ||}
 \hline
 Layer  & KernelSize & Stride & Padding & Activation \\ [0.5ex]
 \hline\hline
 Conv. & 7x7 & 1 & 6 & ReLu \\
 \hline
 Conv. & 3x3 & 1 & 2 & ReLu \\
 \hline
FC & - & - & - & Sigmoid \\
 \hline
 \end{tabular}
\end{minipage}\hfill
}}\vspace*{2mm}
\caption{Simple CNN.}
\label{tab_cnn}
\end{table}
\end{center}

\subsubsection{UNet} 
UNet of \cite{unet} is among the best neural networks for binary segmentation tasks. This architecture consists of two parts: a) encoder that reduces the dimensionality of the original image while increasing the number of channels followed by b) decoder iteratively returning encoded representation to the original image size and channel number. This architecture identifies image features of different scale, improving the quality of the result.

We tune UNet for optimal parameters for our specific problem. A detailed explanation can be found in the \hyperref[sec:app]{Appendix} section.
\begin{enumerate}
    \item Initial number of filters = 16
    \item Kernel size = 5
    \item Number of blocks = 3
\end{enumerate}

The first parameter corresponds to the number of features in the first UNet block. Each subsequent block doubles the number of features. Kernel size is the size of the kernel in the convolution layer.  The \textit{number of blocks} corresponds to the number of blocks in the Decoder and Encoder parts.

We train UNet with EarlyStopping criterion with patience equal to 3 from \cite{earlystopping}. We use Adam optimizer with the learning rate equal to 0.001. Each batch from the training set contains at least half images with no less than one source on it. Due to our image cropping, a batch consists of two parts: one half is random images, the other one contains images with at least one source.

\subsubsection{FoCNN}

Fourier transformation is widely used for denoising tasks in image processing. We combine it with a simple convolutional neural network, which takes several images as an input. Neural networks already contain non-linearity inside, but multiple inputs allow them to provide non-linear information directly. It may reduce network size without efficiency drop.

\begin{figure}[h!]
    \centering
    \includegraphics[width=\columnwidth]{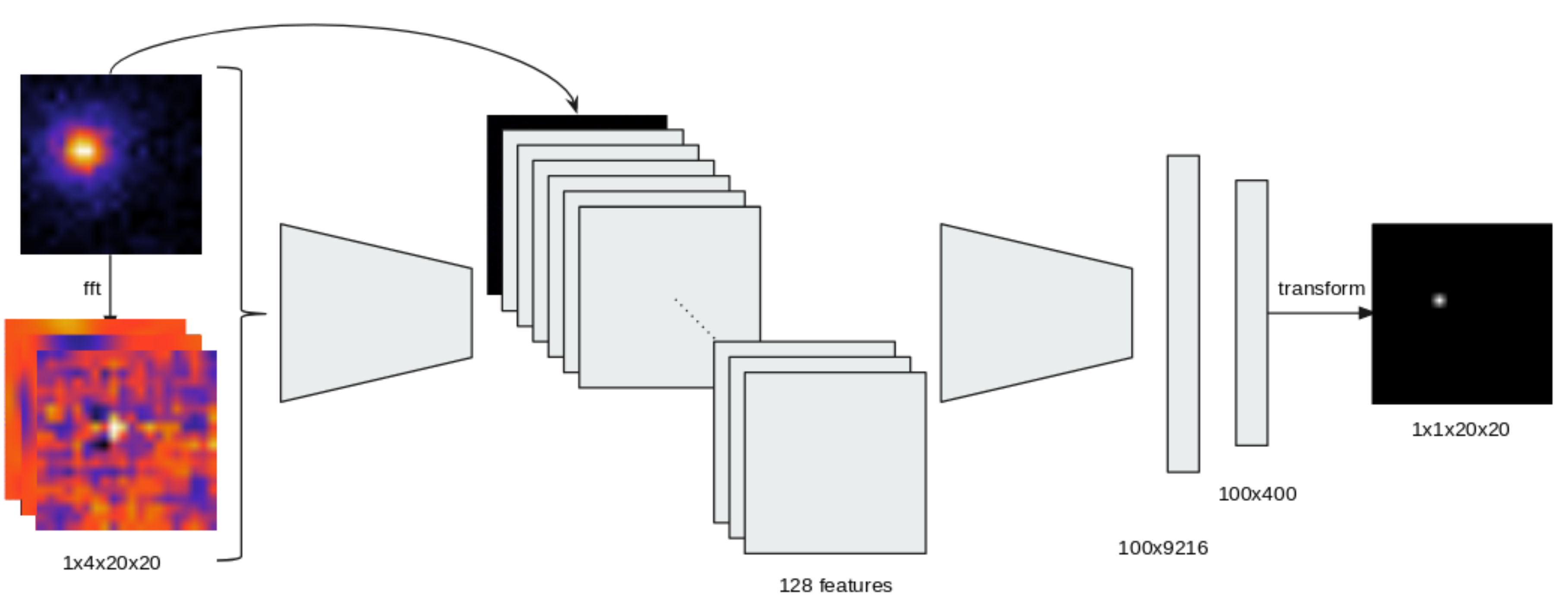}
    \caption{FoCNN Architecture. The network takes a raw count photons image and its transformed versions. One skip connection is added in the middle layers. The output is a map. Its pixels greater than chosen threshold value correspond to predicted point sources. }
    \label{fig:fcnn}
\end{figure}

First, we apply Fourier transformation to the image, then modify a frequency domain image, and finally, return a new image via inverse Fourier transformation. We choose three different bands: 0, 14, and 18 side square cut from the centre of $20 \times 20$ frequency domain image. They are fed to the neural network alongside an unedited image. The architecture can be found in Figure \ref{fig:fcnn}.

After a first convolution block, we concatenate the output with the original astrophysical image to preserve small scale details. We put these features through a second convolution block and two fully-connected layers. The last activation function is a sigmoid. The output is a vector with 400 components where each component corresponds to a pixel of $20 \times 20$ image.

We train this neural network using Adam optimizer for 50 epochs with the linear combination of cross-entropy and MSE loss.

$$\mathcal{L} = \frac{1}{N^2}\sum_{i,j} (y_{i,j} - f(x)_{i,j})^2 + 100*(-\frac{1}{N^2}\sum_{i,j} y_{i,j} \log f(x)_{i,j})$$

\begin{table}[!htb]

      \centering
 \begin{tabular}{|| c | c | c | c | c ||}
 \hline
 Layer  & KernelSize & Stride & Padding & Activ \\ [0.5ex]
 \hline\hline
 Conv. & 3x3 & 1 & 1 & ReLu \\
 \hline
 Conv. & 5x5 & 1 & 2 & ReLu \\
 \hline
 Conv. & 3x3 & 1 & 2 & ReLu \\
 \hline
        \end{tabular}
        
    \bigskip
      \centering
        
 \begin{tabular}{|| c | c | c | c | c ||}
 \hline
 Layer  & KernelSize & Stride & Padding & Activ \\ [0.3ex]
 \hline\hline
 Conv. & 3x3 & 1 & 2 & ReLu \\
 \hline
  MaxPool. & 2x2 & None & 0 & ReLu \\
 \hline
 Conv. & 3x3 & 1 & 2 & ReLu \\
 \hline
   MaxPool. & 2x2 & None & 0 & ReLu \\
 \hline
 Fc. & - & - & - & Sigmoid \\
 \hline
        \end{tabular}
        
    \caption{FoCNN feature extractors.}
\end{table}

\subsection{Postprocessing}
\label{sec:postprocessing}
The output of neural networks are probabilities from 0 to 1. We determine a threshold to predict a source deduced from the results on the validation data. We search for $k$ which gives a threshold $\mu(x) + k*\sigma(x)$ minimizing the metrics score. Here $\mu(x)$ is an average over all pixels and $\sigma(x)$ is a deviation over all pixels across an image $x$. Via optimization on the validation set, we found $k$ equal to 25. So for each picture, we determine its threshold.

To increase precision during the prediction stage, we crop images with intersections(one crop overlaps with another one). We average all predictions across each pixel. Then sources, even of small magnitude are detected with better accuracy. They are detected in multiple crops with a small position change. The neural network predictions from different crops are composed together. Thus, the most probable position of the source will have a more significant score as it is the same place for several crops. This means more precise results for position prediction.

If sources are predicted in two neighbouring pixels, we return only one of them corresponding to the maximum score assigned by the neural network.

\section{Results}
We present the results of the comparison of D3PO, SExtractor, YOLO, CNN, Unet and FoCNN in Table \ref{table:summary}. The traditional method (likelihood-fitting based approach for a point-like source search on top of considered background model) is too computationally expensive since it requires complete simulation of all events and additional analysis. Our implementation of such method gives results worse or comparable to D3PO. Thus we limit ourselves to comparison with D3PO. We plan to compare our approach with the traditional one in the future work which relates to the analysis of the real data.

We characterise the performance of each algorithm by its F1 score, true-positive rate, metrics score, a variation of metrics and testing time for one picture $200 \times 200$ pixels. Metrics score in this table is a mean of all scores in the data set. Variation is also computed over all the data set. We specify that testing time includes only the optimisation step without post-processing. The predicted point is considered true positive if it is contained in a circle with a radius of 1.6 pixels from the real point source which corresponds to the Fermi/LAT characteristic sources localisation accuracy of $0.1^\circ$.

\begin{center}
\begin{table}
\centering
\makebox[0pt][c]{\parbox{0.9\textwidth}
{
\begin{minipage}[b]{1.0\linewidth}\centering
 \begin{tabular}{|| c || c | c | c | c | c ||}
 \hline
 & f1 & TPR  & Mean($\rho$) & Var($\rho$) & $t_\text{test}$ \\[0.5ex]
 \hline\hline
 D3PO & 0.45 & 0.31 & 402 & 227 &  \textgreater $10^6$ ms\\

 \hline
 SExtractor & 0.49 & 0.36 & 346 & 223 & 236 ms\\
 \hline
 CNN & 0.49 & 0.36 & 376 & 217 & {\bf{34 ms}}\\
    \hline
 YOLO & 0.48 & 0.38 & 341 & 185 & 41 ms\\
 \hline
 UNet &  {\bf{0.55}} & 0.38 & 300 & 188 & 44 ms \\
 \hline
 
 FoCNN &  {\bf{0.55}} &  {\bf{0.40}} &  {\bf{298}} &  {\bf{187}} & 51 ms\\
 \hline
\end{tabular}
\end{minipage}\hfill
}}\vspace*{2mm}
\caption{A summary table for test set. F1 and TPR scores are computed for 0.1 degree radius which corresponds to standard Fermi Telescope error (1.6 pixels). If a predicted sources is found within this distance from the original one, the prediction is considered to be correct. $\rho$ is the value of the metrics described in Subsection \ref{desc:metrics}. Mean and Variance are taken over all images from the test set.  $t_\text{test}$ corresponds to an average execution time for one image from the test set.}
\label{table:summary}
\end{table}
\end{center}

\section{Discussion}

As we observe the trade-off between the time and the performance for different algorithms in our result table, we can make the following comments. 

D3PO finds less than half of all point sources, but it rarely detects false sources. Testing time appears to be extremely large due to low convergence rates. 
SExtractor shows better results than D3PO significantly improving mean metrics score over the test set. It takes nearly one-fourth of the second to execute for one test image. Selection of background priors, suggested by the authors, perform poorly on our data set. Its initial parameters give metric scores close to 1200. Thus, we optimize for better priors using a training set. Also, we change the threshold of the algorithm. It improves a bit as the output points tend to cluster around real sources. Thus, to improve the accuracy, one can apply DBSCAN with $\epsilon$=0.01. 

SExtractor algorithm is susceptible to parameters. With default parameters, the metric score is close to 2000.  The optimal settings have been selected as a result of extensive optimization. Though SExtractor can sometimes wrongly predict the positions of the sources, it allows to determine the magnitude (significance) of the point sources which our algorithm does not.


The simple CNN is the first neural network approach tested. Its performance drops comparing to SExtractor while in terms of speed it outperforms the same algorithm by a factor of 6. We notice that SExtractor have the same precision as CNN for f1 and TRP scores. 

YOLO improves true-positive rate and metrics variation while its F1-score is slightly worse compared to two previous approaches. Its metrics score stays close to SExtractor performance with the time execution comparable with CNN.

UNet and FoCNN succeed to improve the results further. Both of them in contrast with previous methods are applied on the cropped images which are combined using the technique described in Sec.~\ref{sec:postprocessing}. Their overall performance is nearly the same, Unet works faster but FoCNN has a better true-positive rate. We also tested them on an extended data set getting 307 and 302 metrics scores respectively. The extended dataset was generated following the same algorithm as the testing set. It contains 900 new samples. 

The significant difference in metrics score for CNN and SExtractor might be the consequence of false positives that are situated closer to real sources in the SExtractor case. Thus, we can conclude that although SExtractor successfully detects a point source in some regions, it fails to localize them correctly.

FoCNN shows better performance on the extended data set both in terms of accuracy and performance. It contained data from the same distribution. It improves accuracy by $\approx 15 \%$ compared to SExtractor and reduces the inference time by a factor of 4.

In real life, there are also extended sources, not only point ones. Our models will most likely miss them interpreting as a part of the diffused background.

\section{Conclusion}
We consider the problem of identification of point source detection for astrophysical images in the gamma range. One of the essential aspects of such identification is independence from prior assumptions on background distribution. We have presented four approaches for point sources extraction based on convolution neural networks and compared them with several baselines methods on a simulated dataset. This dataset is based on real data with a mixture of signal and background, where each image is a raw count photon maps of 10x10 deg$^2$ with energies from 1 to 10 GeV. We demonstrate that our CNN models allow for improving f1-score by $\approx 6 \%$ in comparison to D3PO and SExtractor methods and reducing the inference time by a factor of 4 or 6.

The simulated dataset replicates the main characteristics of the real observations. We argue thus that the CNN models can be generalised to the analysis of the actual astrophysical gamma-ray images.
In our future work, we plan to apply developed algorithms to the real Fermi/LAT data and compare obtained source list to one from 3FGL/4FGL catalogues. Independent analysis will allow to cross-check catalogue results and potentially improve the point sources localisation in background dominated regions in the galactic plane. We also foresee that our study can be modified for a search for a weak extended source characterised by different than background/foreground spectrum. In the last case, the proposed method could be of particular interest for the search and population studies of extended galactic sources (e.g. supernova remnants). A similar approach can also be utilised for searches of the physics beyond the Standard Model, e.g. searches for a weak gamma-ray signal from dark matter annihilation in particular astrophysical objects or dark matter clumps seen in simulations and potentially present in the Milky Way~\citep{acquarius,illustris}.

\noindent\textit{Acknowledgements}
The authors acknowledge support by the state of Baden-W\"urttemberg through bwHPC. The work of D.M. was supported by DFG through the grant MA 7807/2-1.

\bibstyle{Wiley-ASNA}
\bibliography{biblio}

\section{Appendix}
\label{sec:app}



\subsection{UNet}
Optimal parameters for the UNet model were selected from the range shown in Table \ref{table:unet_params}. Also, this table shows the resulted metrics for each configuration.

The largest model shows the lowest loss value for the validation set throughout the learning process. After the end of the training process, the optimal threshold for each model was selected separately, with which they show the best result on the validation set. Resulted metrics shown in Table 4 were calculated on the test set.

Even though the largest model shows the lowest error value during the training process, the best model based on the metric values is UNet with started number of filters=16, the number of blocks=3, kernel size=5, with metrics score 300. Thus we can see that the graph on the validation sample does not correlate with the distance values, it can be concluded that the weighted cross-entropy is not quite the right choice of loss for the given task, the replacement of this loss may likely bring a noticeable improvement in quality.

\begin{center}
\begin{table}
\centering
\makebox[0pt][c]{\parbox{0.9\textwidth}
{
\begin{minipage}[b]{1.0\linewidth}\centering
 \begin{tabular}{|| c || c | c | c | c ||}
 \hline
 UNet Params & F1 Score & TPR & std & Distance\\[0.5ex]
 \hline\hline
 F=8 KS=3 NB=3 & 0.533 & 0.369 & 204 & 316\\
 \hline
 F=8 KS=3 NB=4 & 0.534 & 0.37 & 207 & 318\\
 \hline
 F=8 KS=3 NB=5 & 0.527 & 0.361 & 205 & 319\\
 \hline
 F=8 KS=5 NB=3 & 0.448 & 0.291 & 261 & 394\\
 \hline
 F=8 KS=5 NB=4 & 0.517  & 0.357  & 203  & 328\\
 \hline
 F=8 KS=5 NB=5 & 0.499 & 0.337 & 222 & 348\\
 \hline
 F=16 KS=3 NB=3 & 0.55 & 0.383 & \textbf{188} & 300.8\\
 \hline
 F=16 KS=3 NB=4 & 0.541 & 0.378 & 196 & 308\\
 \hline
 F=16 KS=3 NB=5 & \textbf{0.555} & \textbf{0.39} & 191 & \textbf{300.6}\\
 \hline
 F=16 KS=5 NB=3 & 0.528 & 0.364 & 202 & 317\\
 \hline
 F=16 KS=5 NB=4 & 0.364  & 0.37  & 198 & 310\\
 \hline
 F=16 KS=5 NB=5 & 0.523 & 0.372 & 203 & 317\\
 \hline
 \hline
\end{tabular}
\end{minipage}\hfill
}}\vspace*{2mm}
\caption{The result of UNet model training with different parameters. Here F is a number of filters, KS - kernel size and NB is a number of blocks.}
\label{table:unet_params}
\end{table}
\end{center}

\subsection{FoCNN}

We tested several different architectures of FoCNN as well. A number of layers turned out to be crucial for the performance of the neural network. The architecture presented in Table \ref{focnn1} has a metrics score of 360 that is close to a simple CNN with three layers. Its value does not improve during training computed over the validation dataset.

BatchNormalization added to a final architecture plummets the metrics score to 340.

\begin{table}[!htb]
      
      \centering
 \begin{tabular}{|| c | c | c | c | c ||}
 \hline
 Layer  & KernelSize & Stride & Padding & Activ \\ [0.5ex]
 \hline\hline
 Conv. & 5x5 & 1 & 2 & ReLu \\
 \hline
        \end{tabular}
        
    \bigskip
      \centering
        
 \begin{tabular}{|| c | c | c | c | c ||}
 \hline
 Layer  & KernelSize & Stride & Padding & Activ \\ [0.3ex]
 \hline\hline
 Conv. & 5x5 & 1 & 2 & ReLu \\

 \hline
 Conv. & 3x3 & 1 & 2 & ReLu \\
 \hline
 Fc. & - & - & - & Sigmoid \\
 \hline
        \end{tabular}
    \caption{FoCNN$_{0}$ feature extractors.}
    \label{focnn1}
\end{table}

\input{journals.tex}
\jnlcitation{\cname{%
\author{M. Drozdova}, 
\author{A. Broilovskiy}, 
\author{A. Ustyuzhanin},
\author{D. Malyshev}
} (\cyear{2020}), 
\ctitle{A study of Neural networks point source extraction on simulated Fermi/LAT Telescope images.}, 
\cjournal{ASNA}, \cvol{}.}

\end{document}

%% file: journals.tex
\def\aj{AJ}%
\def\actaa{Acta Astron.}%
\def\araa{ARA\&A}%
\def\apj{ApJ}%
\def\apjl{ApJ}%
\def\apjs{ApJS}%
\def\ao{Appl.~Opt.}%
\def\apss{Ap\&SS}%
\def\aap{A\&A}%
\def\aapr{A\&A~Rev.}%
\def\aaps{A\&AS}%
\def\azh{AZh}%
\def\baas{BAAS}%
\def\bac{Bull. astr. Inst. Czechosl.}%
\def\caa{Chinese Astron. Astrophys.}%
\def\cjaa{Chinese J. Astron. Astrophys.}%
\def\icarus{Icarus}%
\def\jcap{J. Cosmology Astropart. Phys.}%
\def\jrasc{JRASC}%
\def\mnras{MNRAS}%
\def\memras{MmRAS}%
\def\na{New A}%
\def\nar{New A Rev.}%
\def\pasa{PASA}%
\def\pra{Phys.~Rev.~A}%
\def\prb{Phys.~Rev.~B}%
\def\prc{Phys.~Rev.~C}%
\def\prd{Phys.~Rev.~D}%
\def\pre{Phys.~Rev.~E}%
\def\prl{Phys.~Rev.~Lett.}%
\def\pasp{PASP}%
\def\pasj{PASJ}%
\def\qjras{QJRAS}%
\def\rmxaa{Rev. Mexicana Astron. Astrofis.}%
\def\skytel{S\&T}%
\def\solphys{Sol.~Phys.}%
\def\sovast{Soviet~Ast.}%
\def\ssr{Space~Sci.~Rev.}%
\def\zap{ZAp}%
\def\nat{Nature}%
\def\iaucirc{IAU~Circ.}%
\def\aplett{Astrophys.~Lett.}%
\def\apspr{Astrophys.~Space~Phys.~Res.}%
\def\bain{Bull.~Astron.~Inst.~Netherlands}%
\def\fcp{Fund.~Cosmic~Phys.}%
\def\gca{Geochim.~Cosmochim.~Acta}%
\def\grl{Geophys.~Res.~Lett.}%
\def\jcp{J.~Chem.~Phys.}%
\def\jgr{J.~Geophys.~Res.}%
\def\jqsrt{J.~Quant.~Spec.~Radiat.~Transf.}%
\def\memsai{Mem.~Soc.~Astron.~Italiana}%
\def\nphysa{Nucl.~Phys.~A}%
\def\physrep{Phys.~Rep.}%
\def\physscr{Phys.~Scr}%
\def\planss{Planet.~Space~Sci.}%
\def\procspie{Proc.~SPIE}%
\let\astap=\aap
\let\apjlett=\apjl
\let\apjsupp=\apjs
\let\applopt=\ao

%% file: main.bbl
\begin{thebibliography}{}

\bibitem [\protect \citeauthoryear {%
{Ackermann}%
\ \protect \BOthers {.}}{%
{Ackermann}%
\ \protect \BOthers {.}}{%
{\protect \APACyear {2013}}%
}]{%
lat_psf}
\APACinsertmetastar {%
lat_psf}%
\begin{APACrefauthors}%
{Ackermann}, M.%
, {Ajello}, M.%
, {Allafort}, A.%
\ et al.\end{APACrefauthors}%
\unskip\
\newblock
\APACrefYearMonthDay{2013}{Mar}{},
\newblock
\unskip
\newblock
\APACjournalVolNumPages{\apj}{765}{1}{54}.
\newblock
\begin{APACrefDOI} \doi{10.1088/0004-637X/765/1/54} \end{APACrefDOI}
\PrintBackRefs{\CurrentBib}

\bibitem [\protect \citeauthoryear {%
{Ackermann}%
\ \protect \BOthers {.}}{%
{Ackermann}%
\ \protect \BOthers {.}}{%
{\protect \APACyear {2012}}%
}]{%
fermi_galdiffuse}
\APACinsertmetastar {%
fermi_galdiffuse}%
\begin{APACrefauthors}%
{Ackermann}, M.%
, {Ajello}, M.%
, {Atwood}, W\BPBI B.%
\ et al.\end{APACrefauthors}%
\unskip\
\newblock
\APACrefYearMonthDay{2012}{May}{},
\newblock
\unskip
\newblock
\APACjournalVolNumPages{\apj}{750}{1}{3}.
\newblock
\begin{APACrefDOI} \doi{10.1088/0004-637X/750/1/3} \end{APACrefDOI}
\PrintBackRefs{\CurrentBib}

\bibitem [\protect \citeauthoryear {%
{Atwood}%
\ \protect \BOthers {.}}{%
{Atwood}%
\ \protect \BOthers {.}}{%
{\protect \APACyear {2009}}%
}]{%
lat}
\APACinsertmetastar {%
lat}%
\begin{APACrefauthors}%
{Atwood}, W\BPBI B.%
, {Abdo}, A\BPBI A.%
, {Ackermann}, M.%
\ et al.\end{APACrefauthors}%
\unskip\
\newblock
\APACrefYearMonthDay{2009}{Jun}{},
\newblock
\unskip
\newblock
\APACjournalVolNumPages{\apj}{697}{2}{1071-1102}.
\newblock
\begin{APACrefDOI} \doi{10.1088/0004-637X/697/2/1071} \end{APACrefDOI}
\PrintBackRefs{\CurrentBib}

\bibitem [\protect \citeauthoryear {%
Bertin~E.%
}{%
Bertin~E.%
}{%
{\protect \APACyear {1996}}%
}]{%
sxtractor1996}
\APACinsertmetastar {%
sxtractor1996}%
\begin{APACrefauthors}%
Bertin~E., A\BPBI S.%
\end{APACrefauthors}%
\unskip\
\newblock
\APACrefYearMonthDay{1996}{}{},
\newblock
\unskip
\newblock
\APACjournalVolNumPages{Astronomy and Astrophysics Supplement}{117}{}{393-404}.
\PrintBackRefs{\CurrentBib}

\bibitem [\protect \citeauthoryear {%
Fan%
, Su%
\BCBL {}\ \BBA {} Guibas%
}{%
Fan%
\ \protect \BOthers {.}}{%
{\protect \APACyear {2016}}%
}]{%
chamferDistance}
\APACinsertmetastar {%
chamferDistance}%
\begin{APACrefauthors}%
Fan, H.%
, Su, H.%
\BCBL {}\ \BBA {} Guibas, L\BPBI J.%
\end{APACrefauthors}%
\unskip\
\newblock
\APACrefYearMonthDay{2016}{}{},
\newblock
\unskip
\newblock
\APACjournalVolNumPages{CoRR}{abs/1612.00603}{}{}.
\newblock
\begin{APACrefURL} \url{http://arxiv.org/abs/1612.00603} \end{APACrefURL}
\PrintBackRefs{\CurrentBib}

\bibitem [\protect \citeauthoryear {%
Fawcett%
}{%
Fawcett%
}{%
{\protect \APACyear {2006}}%
}]{%
roc}
\APACinsertmetastar {%
roc}%
\begin{APACrefauthors}%
Fawcett, T.%
\end{APACrefauthors}%
\unskip\
\newblock
\APACrefYearMonthDay{2006}{}{},
\newblock
\unskip
\newblock
\APACjournalVolNumPages{Pattern Recognition Letters}{27}{8}{861-874}.
\newblock
\begin{APACrefURL}
  \url{http://dblp.uni-trier.de/db/journals/prl/prl27.html#Fawcett06}
  \end{APACrefURL}
\PrintBackRefs{\CurrentBib}

\bibitem [\protect \citeauthoryear {%
\APACciteatitle {{Fermi Large Area Telescope Fourth Source Catalog}}}{%
\APACciteatitle {{Fermi Large Area Telescope Fourth Source Catalog}}}{%
{\protect \APACyear {2019}}%
}]{%
4fgl_catalogue}
\APACinsertmetastar {%
4fgl_catalogue}%
\unskip
\newblock
\APACrefYearMonthDay{2019}{Feb}{},
\newblock
\APACjournalVolNumPages{arXiv e-prints}{}{}{arXiv:1902.10045}.
\PrintBackRefs{\CurrentBib}

\bibitem [\protect \citeauthoryear {%
\APACciteatitle {{Fermi Large Area Telescope Third Source Catalog}}}{%
\APACciteatitle {{Fermi Large Area Telescope Third Source Catalog}}}{%
{\protect \APACyear {2015}}%
}]{%
3fgl_catalogue}
\APACinsertmetastar {%
3fgl_catalogue}%
\unskip
\newblock
\APACrefYearMonthDay{2015}{Jun}{},
\newblock
\APACjournalVolNumPages{\apjs}{218}{2}{23}.
\newblock
\begin{APACrefDOI} \doi{10.1088/0067-0049/218/2/23} \end{APACrefDOI}
\PrintBackRefs{\CurrentBib}

\bibitem [\protect \citeauthoryear {%
Flamary%
}{%
Flamary%
}{%
{\protect \APACyear {2017}}%
}]{%
fermi_cnn}
\APACinsertmetastar {%
fermi_cnn}%
\begin{APACrefauthors}%
Flamary, R.%
\end{APACrefauthors}%
\unskip\
\newblock
\APACrefYearMonthDay{2017}{}{},
\newblock
\unskip
\newblock
\APACjournalVolNumPages{arXiv:1612.04526v2}{}{}{}.
\PrintBackRefs{\CurrentBib}

\bibitem [\protect \citeauthoryear {%
Goodfellow%
, Bengio%
\BCBL {}\ \BBA {} Courville%
}{%
Goodfellow%
\ \protect \BOthers {.}}{%
{\protect \APACyear {2016}}%
}]{%
Goodfellow2016}
\APACinsertmetastar {%
Goodfellow2016}%
\begin{APACrefauthors}%
Goodfellow, I.%
, Bengio, Y.%
\BCBL {}\ \BBA {} Courville, A.%
\end{APACrefauthors}%
\unskip\
\newblock
\APACrefYear{2016},
\newblock
\APACrefbtitle {Deep Learning} {Deep Learning}.
\newblock
\APACaddressPublisher{}{MIT Press}.
\newblock
\APACrefnote{\url{http://www.deeplearningbook.org}}
\PrintBackRefs{\CurrentBib}

\bibitem [\protect \citeauthoryear {%
H{\"o}gbom%
}{%
H{\"o}gbom%
}{%
{\protect \APACyear {1974}}%
}]{%
clean}
\APACinsertmetastar {%
clean}%
\begin{APACrefauthors}%
H{\"o}gbom, J\BPBI A.%
\end{APACrefauthors}%
\unskip\
\newblock
\APACrefYearMonthDay{1974}{{\APACmonth{06}}}{},
\newblock
\unskip
\newblock
\APACjournalVolNumPages{\aaps}{15}{}{417}.
\PrintBackRefs{\CurrentBib}

\bibitem [\protect \citeauthoryear {%
Krizhevsky%
, Sutskever%
\BCBL {}\ \BBA {} Hinton%
}{%
Krizhevsky%
\ \protect \BOthers {.}}{%
{\protect \APACyear {2012}}%
}]{%
AlexNET}
\APACinsertmetastar {%
AlexNET}%
\begin{APACrefauthors}%
Krizhevsky, A.%
, Sutskever, I.%
\BCBL {}\ \BBA {} Hinton, G\BPBI E.%
\end{APACrefauthors}%
\unskip\
\newblock
\APACrefYearMonthDay{2012}{}{},
\newblock
{\BBOQ}\APACrefatitle {ImageNet Classification with Deep Convolutional Neural
  Networks} {ImageNet Classification with Deep Convolutional Neural
  Networks}.{\BBCQ}
\newblock
\BIn{} F.~Pereira, C\BPBI J\BPBI C.~Burges, L.~Bottou\BCBL {}\ \BOthers {.}\
  (\BEDS), \APACrefbtitle {Advances in Neural Information Processing Systems
  25} {Advances in Neural Information Processing Systems 25}\ \BPGS\
  1097--1105.
\newblock
\APACaddressPublisher{}{Curran Associates, Inc.}
\newblock
\begin{APACrefURL}
  \url{http://papers.nips.cc/paper/4824-imagenet-classification-with-deep-convolutional-neural-networks.pdf}
  \end{APACrefURL}
\PrintBackRefs{\CurrentBib}

\bibitem [\protect \citeauthoryear {%
K.~Schawinski%
}{%
K.~Schawinski%
}{%
{\protect \APACyear {2017}}%
}]{%
gan}
\APACinsertmetastar {%
gan}%
\begin{APACrefauthors}%
K.~Schawinski, C\BPBI Z.%
\end{APACrefauthors}%
\unskip\
\newblock
\APACrefYearMonthDay{2017}{}{},
\newblock
\unskip
\newblock
\APACjournalVolNumPages{Monthly Notices of the Royal Astronomical Society:
  Letters}{467}{}{110-114}.
\PrintBackRefs{\CurrentBib}

\bibitem [\protect \citeauthoryear {%
Marco~Selig%
}{%
Marco~Selig%
}{%
{\protect \APACyear {2015}}%
}]{%
d3po}
\APACinsertmetastar {%
d3po}%
\begin{APACrefauthors}%
Marco~Selig, T\BPBI A.%
\end{APACrefauthors}%
\unskip\
\newblock
\APACrefYearMonthDay{2015}{}{},
\newblock
\unskip
\newblock
\APACjournalVolNumPages{Astronomy and Astrophysics manuscript no. D3PO}{}{}{}.
\PrintBackRefs{\CurrentBib}

\bibitem [\protect \citeauthoryear {%
Marzani%
, Soyez%
\BCBL {}\ \BBA {} Spannowsky%
}{%
Marzani%
\ \protect \BOthers {.}}{%
{\protect \APACyear {2019}}%
}]{%
deepadvertising}
\APACinsertmetastar {%
deepadvertising}%
\begin{APACrefauthors}%
Marzani, S.%
, Soyez, G.%
\BCBL {}\ \BBA {} Spannowsky, M.%
\end{APACrefauthors}%
\unskip\
\newblock
\APACrefYearMonthDay{2019}{}{},
\newblock
\unskip
\newblock
\APACjournalVolNumPages{Lecture Notes in Physics}{}{}{}.
\newblock
\begin{APACrefURL} \url{http://dx.doi.org/10.1007/978-3-030-15709-8}
  \end{APACrefURL}
\newblock
\begin{APACrefDOI} \doi{10.1007/978-3-030-15709-8} \end{APACrefDOI}
\PrintBackRefs{\CurrentBib}

\bibitem [\protect \citeauthoryear {%
NASA.%
}{%
NASA.%
}{%
{\protect \APACyear {2018}}%
}]{%
website:fermi-tools}
\APACinsertmetastar {%
website:fermi-tools}%
\begin{APACrefauthors}%
NASA.%
\end{APACrefauthors}%
\unskip\
\newblock
\APACrefYearMonthDay{2018}{December}{},
\newblock
\APACrefbtitle {Fermi tools software.} {Fermi tools software.}
\newblock
\begin{APACrefURL}
  \url{https://fermi.gsfc.nasa.gov/ssc/data/analysis/software/}
  \end{APACrefURL}
\PrintBackRefs{\CurrentBib}

\bibitem [\protect \citeauthoryear {%
Omohundro%
}{%
Omohundro%
}{%
{\protect \APACyear {1989}}%
}]{%
balltree}
\APACinsertmetastar {%
balltree}%
\begin{APACrefauthors}%
Omohundro, S\BPBI M.%
\end{APACrefauthors}%
\unskip\
\newblock
\APACrefYearMonthDay{1989}{}{},
\newblock
\APACrefbtitle {Five Balltree Construction Algorithms} {Five Balltree
  Construction Algorithms}\ \APACbVolEdTR{}{\BTR{}}.
\PrintBackRefs{\CurrentBib}

\bibitem [\protect \citeauthoryear {%
O.~Ronneberger%
}{%
O.~Ronneberger%
}{%
{\protect \APACyear {2015}}%
}]{%
unet}
\APACinsertmetastar {%
unet}%
\begin{APACrefauthors}%
O.~Ronneberger, T\BPBI B., P.~Fischer.%
\end{APACrefauthors}%
\unskip\
\newblock
\APACrefYearMonthDay{2015}{}{},
\newblock
\unskip
\newblock
\APACjournalVolNumPages{arXiv:1505.04597}{}{}{}.
\PrintBackRefs{\CurrentBib}

\bibitem [\protect \citeauthoryear {%
Pedro~Carvalho%
}{%
Pedro~Carvalho%
}{%
{\protect \APACyear {2011}}%
}]{%
powellsnakes}
\APACinsertmetastar {%
powellsnakes}%
\begin{APACrefauthors}%
Pedro~Carvalho, G\BPBI R.%
\end{APACrefauthors}%
\unskip\
\newblock
\APACrefYearMonthDay{2011}{}{},
\newblock
\unskip
\newblock
\APACjournalVolNumPages{arXiv:1112.4886v1}{}{}{}.
\PrintBackRefs{\CurrentBib}

\bibitem [\protect \citeauthoryear {%
Razzak%
, Naz%
\BCBL {}\ \BBA {} Zaib%
}{%
Razzak%
\ \protect \BOthers {.}}{%
{\protect \APACyear {2017}}%
}]{%
medical_nn}
\APACinsertmetastar {%
medical_nn}%
\begin{APACrefauthors}%
Razzak, M\BPBI I.%
, Naz, S.%
\BCBL {}\ \BBA {} Zaib, A.%
\end{APACrefauthors}%
\unskip\
\newblock
\APACrefYearMonthDay{2017}{}{},
\newblock
\unskip
\newblock
\APACjournalVolNumPages{CoRR}{abs/1704.06825}{}{}.
\newblock
\begin{APACrefURL} \url{http://arxiv.org/abs/1704.06825} \end{APACrefURL}
\PrintBackRefs{\CurrentBib}

\bibitem [\protect \citeauthoryear {%
Redmon%
, Divvala%
, Girshick%
\BCBL {}\ \BBA {} Farhadi%
}{%
Redmon%
\ \protect \BOthers {.}}{%
{\protect \APACyear {2015}}%
}]{%
yolo}
\APACinsertmetastar {%
yolo}%
\begin{APACrefauthors}%
Redmon, J.%
, Divvala, S\BPBI K.%
, Girshick, R\BPBI B.%
\BCBL {}\ \BBA {} Farhadi, A.%
\end{APACrefauthors}%
\unskip\
\newblock
\APACrefYearMonthDay{2015}{}{},
\newblock
\unskip
\newblock
\APACjournalVolNumPages{CoRR}{abs/1506.02640}{}{}.
\newblock
\begin{APACrefURL} \url{http://arxiv.org/abs/1506.02640} \end{APACrefURL}
\PrintBackRefs{\CurrentBib}

\bibitem [\protect \citeauthoryear {%
S.~Pereira%
}{%
S.~Pereira%
}{%
{\protect \APACyear {2016}}%
}]{%
brain}
\APACinsertmetastar {%
brain}%
\begin{APACrefauthors}%
S.~Pereira, V\BPBI A\BPBI C\BPBI A\BPBI S., A.~Pinto.%
\end{APACrefauthors}%
\unskip\
\newblock
\APACrefYearMonthDay{2016}{}{},
\newblock
\unskip
\newblock
\APACjournalVolNumPages{IEEE}{35}{}{1240-1251}.
\PrintBackRefs{\CurrentBib}

\bibitem [\protect \citeauthoryear {%
{Springel}%
\ \protect \BOthers {.}}{%
{Springel}%
\ \protect \BOthers {.}}{%
{\protect \APACyear {2008}}%
}]{%
acquarius}
\APACinsertmetastar {%
acquarius}%
\begin{APACrefauthors}%
{Springel}, V.%
, {Wang}, J.%
, {Vogelsberger}, M.%
\ et al.\end{APACrefauthors}%
\unskip\
\newblock
\APACrefYearMonthDay{2008}{Dec}{},
\newblock
\unskip
\newblock
\APACjournalVolNumPages{\mnras}{391}{4}{1685-1711}.
\newblock
\begin{APACrefDOI} \doi{10.1111/j.1365-2966.2008.14066.x} \end{APACrefDOI}
\PrintBackRefs{\CurrentBib}

\bibitem [\protect \citeauthoryear {%
{Tavani}%
\ \protect \BOthers {.}}{%
{Tavani}%
\ \protect \BOthers {.}}{%
{\protect \APACyear {2009}}%
}]{%
agile}
\APACinsertmetastar {%
agile}%
\begin{APACrefauthors}%
{Tavani}, M.%
, {Barbiellini}, G.%
, {Argan}, A.%
\ et al.\end{APACrefauthors}%
\unskip\
\newblock
\APACrefYearMonthDay{2009}{Aug}{},
\newblock
\unskip
\newblock
\APACjournalVolNumPages{\aap}{502}{3}{995-1013}.
\newblock
\begin{APACrefDOI} \doi{10.1051/0004-6361/200810527} \end{APACrefDOI}
\PrintBackRefs{\CurrentBib}

\bibitem [\protect \citeauthoryear {%
{Vogelsberger}%
\ \protect \BOthers {.}}{%
{Vogelsberger}%
\ \protect \BOthers {.}}{%
{\protect \APACyear {2014}}%
}]{%
illustris}
\APACinsertmetastar {%
illustris}%
\begin{APACrefauthors}%
{Vogelsberger}, M.%
, {Genel}, S.%
, {Springel}, V.%
\ et al.\end{APACrefauthors}%
\unskip\
\newblock
\APACrefYearMonthDay{2014}{Oct}{},
\newblock
\unskip
\newblock
\APACjournalVolNumPages{\mnras}{444}{2}{1518-1547}.
\newblock
\begin{APACrefDOI} \doi{10.1093/mnras/stu1536} \end{APACrefDOI}
\PrintBackRefs{\CurrentBib}

\bibitem [\protect \citeauthoryear {%
Yao%
, Rosasco%
\BCBL {}\ \BBA {} Caponnetto%
}{%
Yao%
\ \protect \BOthers {.}}{%
{\protect \APACyear {2007}}%
}]{%
earlystopping}
\APACinsertmetastar {%
earlystopping}%
\begin{APACrefauthors}%
Yao, Y.%
, Rosasco, L.%
\BCBL {}\ \BBA {} Caponnetto, A.%
\end{APACrefauthors}%
\unskip\
\newblock
\APACrefYearMonthDay{2007}{}{},
\newblock
\unskip
\newblock
\APACjournalVolNumPages{Constructive Approximation}{26}{}{289-315}.
\PrintBackRefs{\CurrentBib}

\bibitem [\protect \citeauthoryear {%
Y.~LeCun%
}{%
Y.~LeCun%
}{%
{\protect \APACyear {1998}}%
}]{%
lenet}
\APACinsertmetastar {%
lenet}%
\begin{APACrefauthors}%
Y.~LeCun, Y\BPBI B., L.~Bottou.%
\end{APACrefauthors}%
\unskip\
\newblock
\APACrefYearMonthDay{1998}{}{},
\newblock
\unskip
\newblock
\APACjournalVolNumPages{Proc. of the IEEE}{}{}{}.
\PrintBackRefs{\CurrentBib}

\end{thebibliography}
